\documentclass{article}

     \usepackage[preprint,nonatbib]{neurips_2019}

\usepackage[utf8]{inputenc} 
\usepackage[T1]{fontenc}    
\usepackage{hyperref}       
\usepackage{url}            
\usepackage{booktabs}       
\usepackage{amsfonts}       
\usepackage{nicefrac}       
\usepackage{microtype}      
\usepackage{amsmath}
\usepackage{amssymb}
\usepackage{graphicx}
\usepackage{todonotes}

\usepackage{soul}
\usepackage{color}

\title{Towards Further Understanding of Sparse Filtering via Information Bottleneck}

\author{%
  Fabio Massimo Zennaro\\
  Department of Informatics,\\
   University of Oslo\\
  PO Box 1080 Blindern, 0316 Oslo, Norway \\
  \texttt{fabiomz@ifi.uio.no} \\
  \And
   Ke Chen \\
   Department of Computer Science,\\
    The University of Manchester \\
   Oxford Rd, Manchester, M13 9PL, UK\\ 
   \texttt{ke.chen@manchester.ac.uk} \\
}

\begin{document}

\maketitle

\section{Introduction (Extended Abstract)}
In this paper we examine a formalization of feature distribution learning (FDL) \cite{ngiam2011sparse,zennaro2018towards} in information-theoretic terms relying on the analytical approach and on the tools already used in the study of the information bottleneck (IB) \cite{tishby2000information}. In \cite{zennaro2018towards} it was suggested that the behavior of FDL algorithms could be expressed as an optimization problem over two information-theoretic quantities, the mutual information of the data with the learned representations and the entropy of the learned distribution. In particular, it was suggested that such a novel formulation might explain the success of the most prominent FDL algorithm, sparse filtering (SF) \cite{ngiam2011sparse}. This conjecture was, however, left unproven. In this work, we aim at providing preliminary empirical support to this conjecture by performing experiments reminiscent of the work done on deep neural networks in the context of the IB research \cite{shwartz2017opening}. Specifically, we borrow the idea of using \emph{information planes} to analyze the behavior of the SF algorithm and gain insights on its dynamics. A confirmation of the conjecture about the dynamics of FDL may provide solid ground to develop information-theoretic tools to assess the quality of the learning process in FDL, and it may be extended to other unsupervised learning algorithms.

\section{Background}
In this section we review the information-theoretic frameworks of interest and state our hypotheses. 

\paragraph{Information Bottleneck Framework.}
In the IB framework \cite{tishby2000information,shwartz2017opening},
(supervised) learning is expressed as a trade-off between (discarding/compressing)
\emph{information in the input data} and (extracting/modelling) \emph{information in the output data}:
\[
\min_{p(t),p(t\vert x),p(y\vert t)}I\left[X;T\right]-\beta I\left[T;Y\right],
\]
where $I[\cdot;\cdot]$ is the mutual information, $\beta$ is a trade-off term, $X$ is the original input data, $T$ is the learned (intermediate) representation, and $Y$ is the output (label) data.

\paragraph{Feature Distribution Learning Framework.}
A formalization of FDL in information-theoretic terms is offered in Equation 1 in \cite{zennaro2018towards}; in this work, it is conjectured that unsupervised FDL can be expressed as a trade-off between (retaining) \emph{information in the input data} and (maximizing) \emph{information in the learned representation}:
\begin{align}
\max_{p(T)\in\mathfrak{P}} & D_{KL}\left[p\left(X,T\right)\parallel p(X)p(T)\right]+D_{KL}\left[p(T)\parallel q(T)\right]\\
\max_{p(T)\in\mathfrak{P}} & I\left[X;T\right]+H[T]
\end{align}
where $D_{KL}\left[\cdot\parallel\cdot\right]$ is the KL divergence, $H[\cdot]$ is Shannon entropy, $p(X)$ is the original data distribution, $p(T)$ is the learned distribution, and $q(T)$ is an entropy-maximizing distribution over the domain of the learned representations. The passage from Equation (1) to Equation (2) relies on simple information-theoretic definitions \cite{cover2012elements}.

\paragraph{Connection between the frameworks.}
The two frameworks have interesting analogies, in that they both express the learning objective as an optimization problem over information-theoretic terms.

Concerning the first term, both frameworks consider the same quantity: $I\left[X;T\right]$. What distinguishes the two approaches is the direction of optimization. In the IB framework $I\left[X;T\right]$ is minimized because in a supervised setting it is desirable to get rid of information that is not related with the task at hand. In the FDL framework $I\left[X;T\right]$ is maximized because in an unsupervised setting there is no task that can be used to discriminate what information is relevant and what is not. This maximization of the mutual information corresponds to the \emph{infomax principle} \cite{linsker1989application}.

Concerning the second term, the difference between the two frameworks is radical, caused by the different setting in which they operate. In the supervised setting, the IB framework exploits label information to discover relevant information and to preserve it. In the FDL framework, no label information is available, and the new objective is to generate representations that may be highly informative in themselves by maximizing the entropy $H[T]$ of the learned representations.

\paragraph{Sparse Filtering.}
SF is an unsupervised learning algorithm that transforms data $X$ as 
\[
T=\ell_{2,col} \left( \ell_{2,row} \left(  \left| WX \right| \right)   \right),
\]
where $W$ is a weight matrix, $\left|\cdot\right|$ is the element-wise absolute-value function, $\ell_{2,row}$ is the $\ell_2$-normalization along the rows, and $\ell_{2,col}$ is the $\ell_2$-normalization along the columns.
The weights $W$ are learned with the simple objective of minimizing the $\ell_1$-sparsity of the learned representation, within the constraints given by the $\ell_2$-normalizations \cite{zennaro2018towards}.
Notice that a SF module can be seen as a shallow neural network with a non-linearity given by $\ell_{2,col} \left( \ell_{2,row} \left(  \left| \cdot \right| \right) \right)$.

\paragraph{Information plane.}
In the study of IB, the behavior of deep neural networks has been studied analyzing their trajectories on the \emph{information plane} \cite{shwartz2017opening}, that is a Cartesian plane where the x-axis is $I\left[X;T\right]$ and the y-axis is $I\left[T;Y\right]$.
Since an algorithm such as SF may be seen as a neural network, we suggest that its behavior may be analyzed by studying its trajectories on an analogous information plane, where the x-axis is again $I\left[X;T\right]$, but the y-axis is now $H[T]$.   

\paragraph{Hypotheses.}
Relying on the estimation of information-theoretic quantities and information plane graphs, we aim at empirically validate the conjecture that the behavior of FDL algorithms represented by SF comply with the information-theoretic description offered in Equation (1). To do so, we try to falsify the conjecture by analyzing the simulated behavior of SF and positing the following negative questions: (i) \emph{can we empirically validate that SF does not work by maximizing} $I\left[X;T\right]$? (ii) \emph{can we empirically validate that SF does not work by maximizing} $H[Z]$? A failure in disproving experimentally the conjecture behind Equation (1) would not prove the conjecture itself, but provide solid evidence about its validity.

\section{Simulations}

In this section we run simulations\footnote{The code for all our simulations is available at \url{https://github.com/FMZennaro/SF-IB}} of the SF algorithm\footnote{The code for the SF algorithm is available at \url{https://github.com/jngiam/sparseFiltering} and \url{https://github.com/FMZennaro/PSF}}, and we compute the information-theoretic quantities of interest\footnote{We estimate information-theoretic quantities using the binning algorithm used in \cite{shwartz2017opening}, available at \url{https://github.com/ravidziv/IDNNs}, and reproduced in \cite{saxe2018information}, available at \url{https://github.com/artemyk/ibsgd}} using a binning algorithm with $30$ bins. We generate four different data sets with increasing complexity: (1) a simple univariate data set replicating the synthetic data from \cite{zennaro2018towards}; (2) a multivariate data set mapping to a lower dimensional representation; (3) a multivariate data set mapping to a higher dimensional representation; (4) a data set with a dimensionality approaching the input data used in \cite{shwartz2017opening}. These simulations allow us to verify the behavior of SF for varied cases within a dimensionality that allows for an effective computation of information-theoretic quantities. We always train on $900$ samples and compute our statistics on $100$ test samples. All simulations are repeated $10$ times.

\paragraph{Simulation 1 - Univariate Low-Dimensional Data}
We generate 2D data sampling two independent Gaussian $\mathcal{N}\left(\mu=0,\sigma=.5\right)$, and we train a SF module to project them in a 2D space.

\paragraph{Simulation 2 - Multivariate Low-Dimensional Data to a Lower Space}
We generate 4D data sampled from a multivariate Gaussian with a mean vector whose entries are sampled from a uniform pdf $\mathcal{U}\left(a=-5,b=5\right)$ and with a random symmetric positive-definite matrix as a covariance matrix. We train a SF module to project the data into a lower 2D space.

\paragraph{Simulation 3 - Multivariate Low-Dimensional Data to a Higher Space}
We generate data as in the previous simulation, but we now train a SF module to project the data into a higher 8D space.

\paragraph{Simulation 4 - Multivariate Data}
We generate 10D data sampled from a multivariate Gaussian as in Simulation 2. We train a SF module to project the data into a lower 4D space.

\begin{figure}
	\centering
	\includegraphics[scale=0.55]{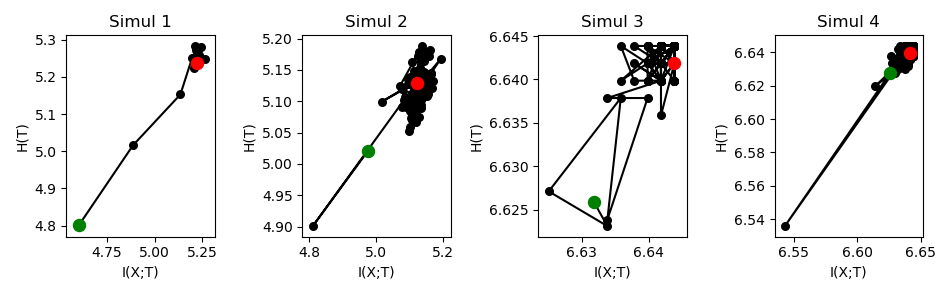}
	\caption{Information graphs for Simulations 1 to 4. Each dot represents the a value of mutual information $I[X;T]$ and entropy $H[T]$ at a given iteration; the green dot denotes the starting value, while the red dot denotes the value at the end of training. The position of each dot is computed as an average of the position over all the simulations. Since each simulation may last a different number of iterations $N_i$, our averages are computed until one simulation stops, that is, up to $\min_i N_i$.} \label{img:IG1}
\end{figure}

\paragraph{Results.}
Figure \ref{img:IG1} reports the information graph of all our simulations. In all the cases, the dynamics shows a progression towards learning representations that have higher mutual information $I[X;T]$ and higher entropy $H[T]$ than the initial state. 
The aggregate dynamics are confirmed by the dynamics of individual simulations, which are available in Appendix \ref{app:SuppFigs}. Some specific behaviors may spur further investigation; for instance, in Simulation 2, dramatic changes in the weights late in the training (and associated with a drop in mutual information and entropy) raise questions about such a transition; in Simulation 3, discrete variations in the information plane make us wonder about the proper choice of binning. A closer look to stark transitions in the learning process and a sensitivity analysis in the computation of information-theoretic quantities are obvious directions of further study.

\section{Discussion}

This paper shows an application of the principle and tools from the IB theory to the study of a FDL  algorithm. Even if this analysis is preliminary, it still allows us to get confirmation of the conjecture behind Equation (1) and obtain insights in the dynamics of learning of SF that may be object of further study. 

In all our simulations we observed an increase in mutual information $I[X;T]$ and entropy $H[T]$ during learning; while this does not allow us to conclude that, in general, FDL algorithms always maximize this quantities, it strongly suggests that SF may work as an effective proxy for this maximization.
  
A finer-grained understanding of the dynamics of SF may also be gotten by using a training procedure based on standard batch gradient descent. Current implementation solves the optimization problem of SF iteratively using L-BFGS-B given the analytical gradient and the whole training data.

The IB approach to SF may also allow to answer more interesting questions. In \cite{zennaro2018towards} it was proved that SF is guaranteed to learn useful representations when the distribution $p(Y\vert X)$ can be explained by a cosine metric; relying on labeled data it would be interesting to find confirmation or denial of this property by analyzing on an actual information plane the dynamics of the the mutual information between the learned representations and the labels.
Finally, the IB theory may provide a framework to answer a question about stacking; while advocated in the original paper \cite{ngiam2011sparse}, stacking of SF modules has not led to any successful application in practice. It may be argued and experimentally studied whether the composition of SF modules in a Markov chain poses too serious a challenge for the objective of maximizing the mutual information between learned representations an the input.

\bibliographystyle{plain}
\bibliography{../../lib/lib}

\appendix

\newpage

\section{Supporting Figures}\label{app:SuppFigs}

\begin{figure}[h]
	\centering
	\includegraphics[scale=0.55]{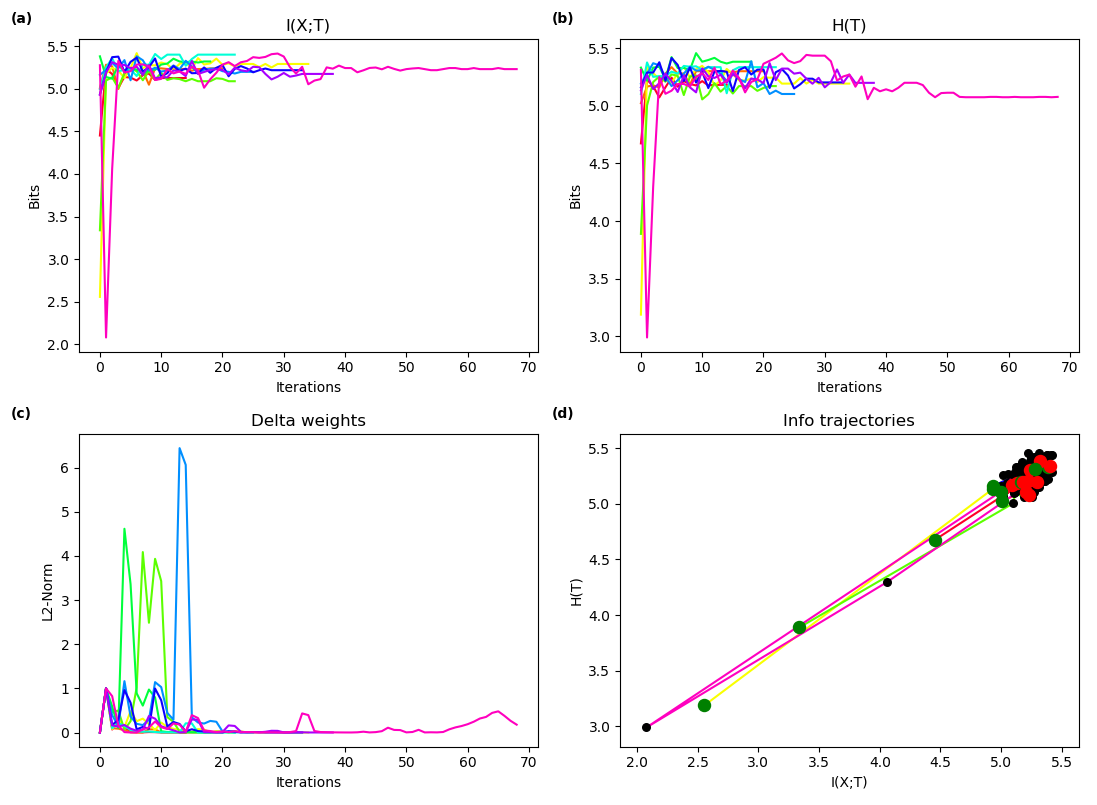}
	\caption{Training dynamics of Simulation 1. Each color represents one of the ten simulations run. (a) Plot of the variation of the mutual information $I[X;T]$ as a function of the iterations. (b) Plot of the variation of the entropy $H(T)$ as a function of the iterations. (c) Plot of the variation in the weights computed using the $L2$-norm as a function of the iterations. (d) Plot of the information graph. }
\end{figure}

\begin{figure}[h]
	\centering
	\includegraphics[scale=0.55]{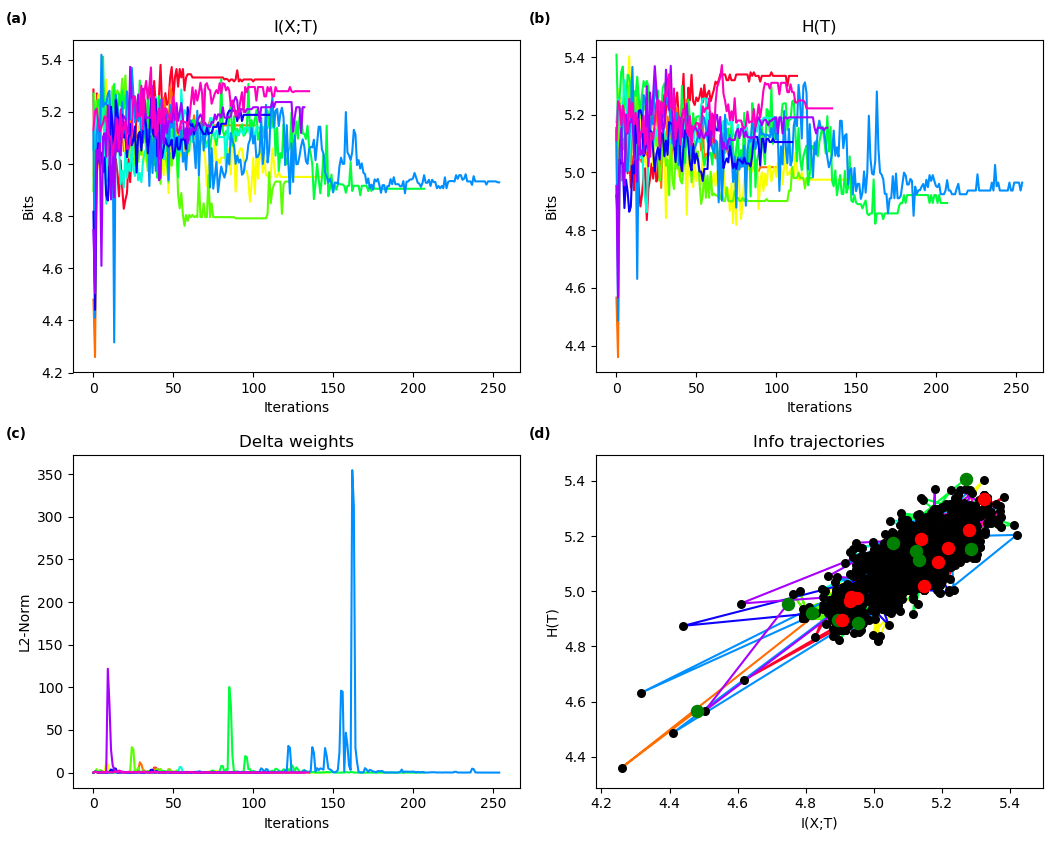}
	\caption{Training dynamics of Simulation 2. Each color represents one of the ten simulations run. (a) Plot of the variation of the mutual information $I[X;T]$ as a function of the iterations. (b) Plot of the variation of the entropy $H(T)$ as a function of the iterations. (c) Plot of the variation in the weights computed using the $L2$-norm as a function of the iterations. (d) Plot of the information graph. }
\end{figure}

\begin{figure}[h]
	\centering
	\includegraphics[scale=0.55]{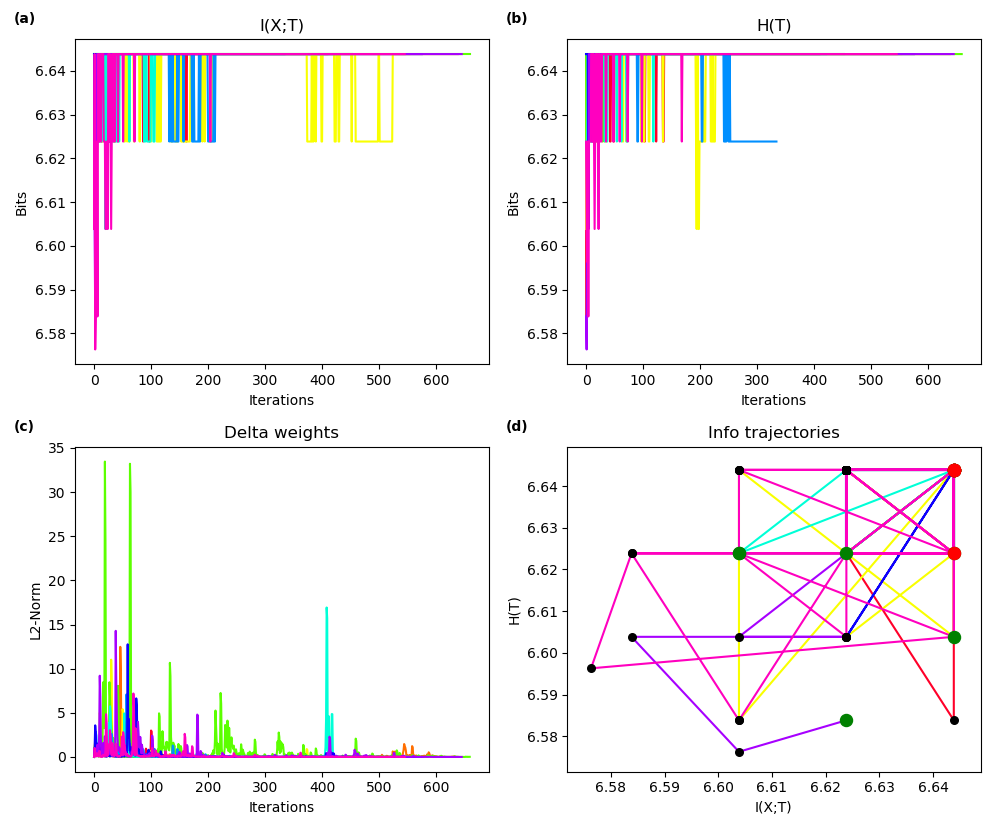}
	\caption{Training dynamics of Simulation 3. Each color represents one of the ten simulations run. (a) Plot of the variation of the mutual information $I[X;T]$ as a function of the iterations. (b) Plot of the variation of the entropy $H(T)$ as a function of the iterations. (c) Plot of the variation in the weights computed using the $L2$-norm as a function of the iterations. (d) Plot of the information graph. }
\end{figure}

\begin{figure}[h]
	\centering
	\includegraphics[scale=0.55]{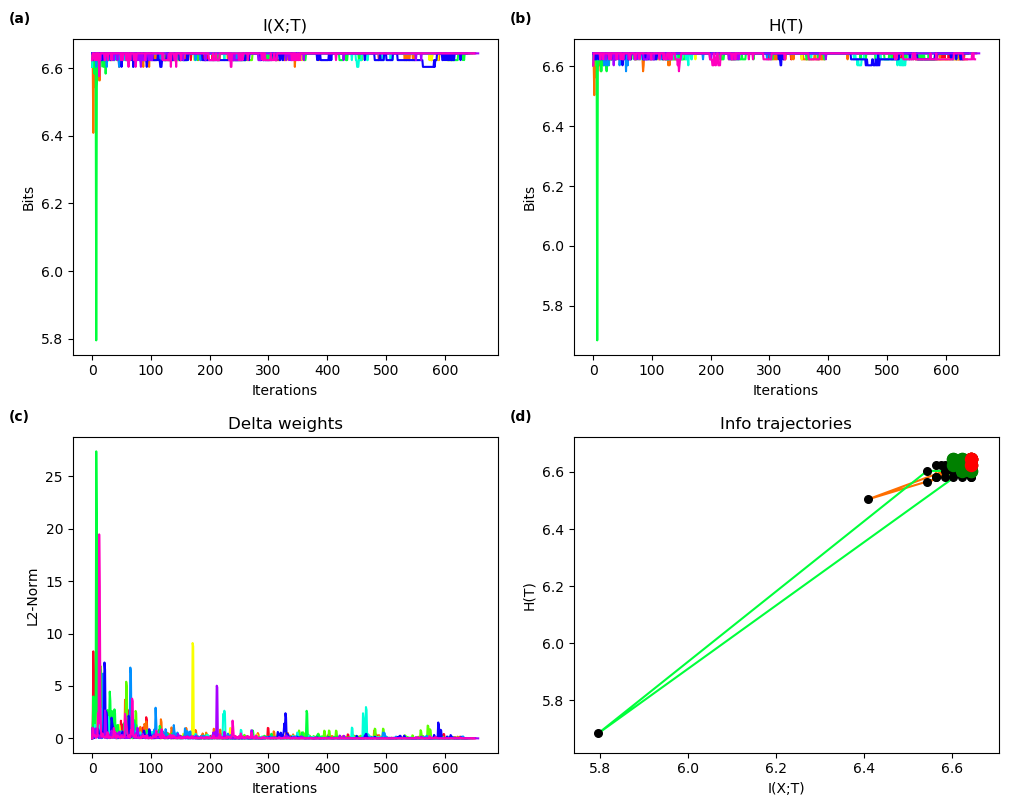}
	\caption{Training dynamics of Simulation 4. Each color represents one of the ten simulations run. (a) Plot of the variation of the mutual information $I[X;T]$ as a function of the iterations. (b) Plot of the variation of the entropy $H(T)$ as a function of the iterations. (c) Plot of the variation in the weights computed using the $L2$-norm as a function of the iterations. (d) Plot of the information graph. }
\end{figure}

\end{document}